\documentclass[conference]{IEEEtran}
\ifCLASSINFOpdf
  \usepackage[pdftex]{graphicx}
  \graphicspath{{../pdf/}{../jpeg/}}
  \DeclareGraphicsExtensions{.pdf,.jpeg,.png}
\else
  \usepackage[dvips]{graphicx}
  \graphicspath{{../eps/}}
  \DeclareGraphicsExtensions{.eps}
\fi
\usepackage{graphicx}

\usepackage{algorithmic}

\usepackage{array}

\usepackage{eqparbox}

\usepackage[tight,footnotesize]{subfigure}
\usepackage{hyperref}

\usepackage[table]{xcolor}
\usepackage{graphicx}
\usepackage[english,ruled,vlined]{algorithm2e}
\SetKwFor{Para}{para}{fa�a}{fim para}
\SetKwRepeat{Do}{do}{while}
\usepackage{algorithmic}
\usepackage{adjustbox}
\usepackage{setspace}
\usepackage{url}
\usepackage{amsmath}
\usepackage{amsthm}
\usepackage{mathrsfs}
\usepackage{mathptmx}
\usepackage{amssymb}
\usepackage{bbm}
\usepackage{verbatim}
\usepackage{calc}
\usepackage{multirow}
\usepackage{epstopdf}
\usepackage{cite}
\usepackage{graphicx}
\usepackage{caption}
\usepackage{hyperref}
\usepackage{enumerate}
\usepackage{amssymb}
\usepackage{memhfixc}
\usepackage{listings}
\usepackage{marvosym}
\usepackage{pgf}
\usepackage{float}
\usepackage{indentfirst}
\usepackage{verbatim}
\usepackage{tikz}
\usetikzlibrary{arrows,automata,positioning,trees,shapes}
\usepackage{mdframed}
\usepackage{tabularx}
\newtheorem{definition}{Definition}

\begin{document}
%
\title{Counterexample Guided Inductive Optimization Applied to Mobile Robots Path Planning \\ (Extended Version)}

%
%
%


\author{
	\IEEEauthorblockN{Rodrigo F. Ara\'ujo\IEEEauthorrefmark{1}, Alexandre Ribeiro\IEEEauthorrefmark{2}, Iury V. Bessa\IEEEauthorrefmark{2}, Lucas C. Cordeiro\IEEEauthorrefmark{3}\IEEEauthorrefmark{2}, Jo\~ao E. C. Filho\IEEEauthorrefmark{2}}
	\IEEEauthorblockA{\IEEEauthorrefmark{1}Federal Institute of Amazonas, Brazil}
	\IEEEauthorblockA{\IEEEauthorrefmark{2}Federal University of Amazonas, Brazil}
	\IEEEauthorblockA{\IEEEauthorrefmark{3}University of Oxford, UK}
}

\markboth{IEEE Latin America Transactions,~Vol.~6, No.~1, January~2007}%
{Shell \MakeLowercase{\textit{et al.}}: Bare Demo of IEEEtran.cls for Journals}
%
\maketitle

\begin{abstract}
We describe and evaluate a novel optimization-based off-line path planning algorithm for mobile robots based on the Counterexample-Guided Inductive Optimization (CEGIO) technique. CEGIO iteratively employs counterexamples generated from Boolean Satisfiability (SAT) and Satisfiability Modulo Theories (SMT) solvers, in order to guide the optimization process and to ensure global optimization. This paper marks the first application of those solvers for planning mobile robot path. In particular, CEGIO has been successfully applied to obtain optimal two-dimensional paths for autonomous mobile robots using off-the-shelf SAT and SMT solvers.
\end{abstract}


%
\IEEEpeerreviewmaketitle

\section{Introduction}
\label{sec:intro}

Recently, mobile robots have been employed for various tasks, replacing humans in dangerous and monotonous tasks with a certain degree of efficiency and safety~\cite{Karma2015,Varela2014,Naidoo2011}. For this specific reason, robot mobile navigation, and in particular, path planning has become an important research topic in recent days.
The basic path planning problem can be described as the computation of robot positions and motions, which allow the robot to autonomously move from one starting point to a final desired position, performing a specific task and avoiding possible obstacles~\cite{Lunenburg2016}. 

There are several methods in the literature that are usually employed for path planning \cite{Bakdi201795,Topalov1997351,DaSilvaArantes2017,BABEL2014142,Han201735,LEKKAS20161,Zeng20161509}. For instance, Yang {\it et al.}~\cite{Yang2016} classify the path planning method in five main categories: sampling-based, node-based optimal, mathematical model-based, bioinspired, and multifusion-based algorithms. Additionally, path planning strategies can be classified in two categories: on-line and off-line. In on-line mode, the path planning can be performed during the robot movement, while in off-line mode, the robot movement path planning is performed in advance ({\it i.e.}, before the movement).

Path planning task is often modeled as an optimization problem, where a decision variable represents a given path, {\it i.e.}, the sequence of points (or movements) by which the robot must move; the cost function is a certain criteria or metric whose value is optimized ({\it e.g.}, distance, energy consumption, and execution time). Thus, various optimization techniques have been applied to solve path planning problems, {\it e.g.}, genetic algorithm (GA)~\cite{Bakdi201795,Topalov1997351,DaSilvaArantes2017}, A*~\cite{BABEL2014142}, particle swarm optimization (PSO)~\cite{Han201735}, nonlinear programming (NLP)~\cite{LEKKAS20161}, and ant colony~\cite{Zeng20161509}. However, these optimization techniques are unable to ensure the global optimality of the robot path, although they are able to provide results sufficiently fast for on-line path planning applications.

We describe and evaluate a novel off-line path planning algorithm based on the counterexample guided inductive optimization (CEGIO) technique described by Araujo {\it et al.}~\cite{Araujo2016,Araujo2017}, which is a Boolean Satisfiability (SAT) and Satisfiability Modulo Theories (SMT) based optimization algorithm that executes iteratively to achieve global optimization via counterexamples produced by SAT and SMT solvers. Previous studies~\cite{Araujo2016,Araujo2017}, showed that CEGIO is able to ensure the global optimization of various non-trivial functions classes ({\it e.g.}, convex, nonlinear, and nonlinear functions), with an accuracy rate better than traditional optimization techniques ({\it e.g.}, GA, PSO, and NLP). Therefore, the main original contributions of this paper are: 

\begin{itemize}
  \item{Apply the CEGIO algorithm to obtain optimal two-dimensional paths for autonomous mobile robots.}
  \item{Evaluate the effect of the minimum improvement (step) on the cost function at each iteration of the CEGIO algorithm.}
\end{itemize}

{\it Outline}: Section~\ref{sec:back} provides a brief background about optimization problems, path planning for mobile robots, software model checking, and CEGIO. Section~\ref{sec:method} describes the proposed methodology steps for efficiently solving optimization problems using the CEGIO approach. Section~\ref{sec:exps} presents the objectives and experimental results of the proposed algorithm if applied to the path planning problem in bi-dimensional space for mobile robots. Finally, Section~\ref{sec:conc} presents the conclusions of this work and outlines future studies.

\section{Background}
\label{sec:back}

\subsection{Optimization Problems}
\label{ssec:opt}
Let $f:X\rightarrow\mathbb{R}$ be a cost function, such that $X \in R^n$ represents the decision variables vector $x_1, x_2, ..., x_n$ and $f(x_1, x_2, ..., x_n) \equiv f(x)$. Let $\Omega \subset X $ be a subset settled by a set of constraints.

\begin{definition}
\label{def:def1}
A multi-variable optimization problem consists in finding an optimal vector $x$, which minimizes $f$ in $\Omega$.
\end{definition}

According to Definition \ref{def:def1}, an optimization problem can be written as
\begin{equation}
\label{eq:optproblem}
\begin{array}{cc}
\min\limits_{x} & f(x),  \\
\textrm{ s.t. } & x \in \Omega. \\
\end{array}
\end{equation}

In particular, this optimization problem can be classified in different ways with respect to constraints, decision variables domain, and nature of cost function $f$. All optimization problems considered here are constrained, \textit{i.e.}, decision variables are constrained by the subset $\Omega$. The optimization problem domain $X$ that contains $\Omega$ can be the set of $\mathbb{N}$, $\mathbb{Z}$, $\mathbb{Q}$, or $\mathbb{R}$. Depending on the domain and constraints, the optimization search-space can be small or large, which influence the optimization algorithm performance.

The cost function can be classified as linear or non-linear; continuous, discontinuous or discrete; convex or non-convex. Depending on the cost function nature, the optimization problem can be difficult to solve, given the time and memory constraints~\cite{GALPERIN19911}, even if we use SAT and SMT solvers~\cite{DBLP:journals/dafes/TrindadeC16}. Particularly, non-convex optimization problems are the most difficult ones with respect to the cost function nature. A non-convex cost function is a function whose epigraph is a non-convex set and consequently presents various inflexion points that can trap the optimization algorithm to a sub-optimal solution. A non-convex problem is necessarily a non-linear problem and it can also be discontinuous. Depending on that classification, some optimization techniques are unable to solve the optimization problem, and some algorithms usually point to suboptimal solutions, \textit{i.e.}, a solution that is not a global minimum of $f$, but it only locally minimizes $f$. Global optimal solutions of the function $f$, aforementioned, can be defined as:
\begin{definition}
	\label{def:def2}
	A vector $x^* \in \Omega$ ia a global optimal solution of $f$ in $\Omega$ iff $f(x^*)\leq f(x)$, $\forall x \in \Omega$.
\end{definition}
\subsection{Path Planning for Mobile Robots}
\label{ssec:path}

Path planning is one of the main robot navigation steps in which the robot must determine a safe and a collision free path from one starting point to a target point. Such path is a curve without time continuous consideration and it is composed of various segments, usually called by trajectories~\cite{Yang2016}. 

\begin{definition}
	\label{def:path}
	A path is a set of straight segments successively connected to guide the mobile robot from one initial point to a target point.
\end{definition}

A path planning algorithm can be evaluated, given some properties, {\it e.g.}, completeness, optimally, correctness, robustness, and computational complexity \cite{Lunenburg2016}. In particular, path planning algorithms often seek to obtain not only a correct path, {\it i.e.}, a safe path that meets the path specification ({\it e.g.}, obstacle avoidance), but also an optimal path~\cite{Yang2016}.

\begin{definition}
	\label{def:optpath}
	An optimal path is a set of straight segments successively connected to guide the mobile robot from one initial point to a target point, which minimizes a cost function related to that path.
\end{definition}

Thus, the main path planning algorithm objective is to find the $n$ points, which form the path that connects the initial and target points of the mission, generating smallest displacement cost, \textit{i.e.}, the path planning is essentially a trajectory optimization problem. In the present work, this problem is restricted to the bi-dimensional environment. 

\subsection{Model Checking}
\label{sssec:mdlchecking}
Model checking is an automated verification procedure to exhaustively check all (reachable) system's states \cite{DBLP:books/daglib/0020348}. The model checking procedure typically consists of three steps: modeling, specification, and verification.

Modeling is the first step, where it converts the system to a formalism that is accepted by a verifier. The second step is the specification, which describes the system's behavior and the property to be checked. Model checking provides ways to check whether a given specification satisfies a system property, but it is difficult to determine whether such specification covers all properties in which the system should satisfy.
Finally, the verification step checks whether a given property is satisfied with respect to a given model, {\it i.e.}, all relevant system states are checked to search for any state that violates the verified property. In case of a property violation, the verifier reports the system execution trace (counterexample), which contains all steps from the initial state to the bad state that leads to the property violation.

\subsubsection{Bounded Model Checking (BMC)}
\label{sssec:BMC}
BMC is an important verification technique, which is based on SAT~\cite{ABiere2009} or SMT~\cite{Barret2009} solvers. BMC has been successfully applied to verify single- and multi-threaded programs~\cite{Cordeiro,DBLP:journals/sttt/GadelhaIC17,DBLP:journals/stvr/MonteiroGCF17,DBLP:phd/ethos/Cordeiro11}. It checks the negation of a given property at a given depth over a transition system $M$.
\begin{definition}
	\label{def:def3}
	Given a transition system $M$, a property $\phi$, and a bound $k$; BMC unrolls the system $k$ times and translates it into a verification condition (VC) $\psi$, which is satisfiable iff $\phi$ has a counterexample of depth less than or equal to $k$~\cite{ABiere2009}.
\end{definition}

In BMC, the associated problem is formulated by constructing the following logical formula
\begin{equation}
\label{eq:logiceq}
\psi_k = I(S_0) \wedge
\bigvee_{i=0}^k
  \bigwedge_{j=0}^{i-1}
   \left(
    \gamma(s_{j},s_{j+1}) \wedge \neg \phi(s_i)
   \right),
\end{equation}

\noindent where $\phi$ is a property and $S_0$ is a set of initial states of $M$, and $\gamma(s_j, s_{j+1})$ is the transition relation of $M$ between time steps $j$ and $j + 1$. Hence, $I(S_{0}) \wedge \bigwedge_{j=0}^{i-1} \gamma (s_{j},s_{j+1}) $ represents the executions of a transition system $M$ of length $i$. The above VC $\psi_k$ can be satisfied if and only if, for some $i \leq k$ there exists a reachable state at time step $i$ in which $\phi$ is violated. If the logical formula~\eqref{eq:logiceq} is satisfiable (\textit{i.e.}, returns true), then the SMT solver provides a satisfying assignment (counterexample).

\begin{definition}
	\label{def:def4}
	A counterexample for a property $\phi$ is a sequence of states $s_0, s_1, . . . , s_k$ with $s_0 \in S_0$, $s_k \in S_k$, and $\gamma (s_i, s_{i+1})$ for $0 \leq i \leq k$ that makes Eq.~\eqref{eq:logiceq} satisfiable. If it is unsatisfiable (\textit{i.e.}, returns false), then we can conclude that there is no error state in $k$ steps or less.
\end{definition}

\subsection{Counterexample Guided Inductive Optimization}
\label{ssec:cegio}
This section presents a novel class of search-based optimization algorithm that employs non-deterministic representation of decision variables and constrains the state-space search based on counterexamples produced by a SAT or SMT solver, in order to ensure the complete global optimization without employing randomness. This class of techniques is defined here as counterexample guided inductive optimization (CEGIO), which is inspired by the syntax-guided synthesis (SyGuS) to perform inductive generalization based on counterexamples provided by a verification oracle \cite{Alur2013}.

In particular, CEGIO relies on iterative executions to constrain a verification procedure, in order to perform inductive generalization, based on counterexamples extracted from SAT and SMT solvers. CEGIO is able to successfully optimize a wide range of functions, including non-linear and non-convex optimization problems based on SAT and SMT solvers, in which data provided by counterexamples are employed to guide the verification engine, thus reducing the optimization domain \cite{Araujo2016}. The function evaluation and the search for the optimal
solution are performed by means of an iterative execution of successive verifications based on counterexamples extracted from SAT and SMT solvers. The counterexample provides new domain boundaries and new optimal candidates. In contrast to other heuristic methods ({\it e.g.}, genetic algorithms), which are usually employed for optimizing this class of function, the present approach always finds the global optimal point.

\section{CEGIO-based Path Planning}
\label{sec:method}

In this section, a novel optimization method to solve the path planning problem of autonomous mobile robots is described. The main objective of a path planning algorithm is to generate points needed to guide the mobile robot in a environment with obstacles.
As observed by other researchers, it is a challenge to find the optimal path in a region, since the number of activities and obstacles types in the environment can substantially increase the path planning problem complexity~\cite{Lunenburg2016}. In most cases, it requires substantial processing time, especially if there are many points to visit.
Therefore, there is a growing need for developing techniques for optimal path planning, considering trajectory length and system energy consumption.

For the path planning method proposed here, the following two steps are applied: (1) encode the environment, movement space, and static obstacles ({\it i.e.}, the environment is assumed to be known and contains only static obstacles); (2) use a path search method that consists of points in the space and its respective orientations, to find a path that satisfies the constraints given by the problem.

Ara\'{u}jo {\it et al.} \cite{Araujo2017} proposed three different algorithm types based on CEGIO, which are suitable for different situations: the Generalized Algorithm (CEGIO-G), the Simplified Algorithm (CEGIO-S), and the Fast Algorithm (CEGIO-F). CEGIO-G can to be applied to any function class ({\it i.e.}, convex and non-convex ones). CEGIO-S is suitable for functions about which we have some prior knowledge (\textit{e.g.}, semi- and positive-definite functions). Finally, CEGIO-F can be applied to convex functions and uses its properties to restrict the associated state-space, according to the results presented by Ara\'{u}jo {\it et al.} \cite{Araujo2017}, which show considerable improvement regarding optimization times. The proposed method consists of modeling the path planning as an optimization problem and solving it by a CEGIO-F based algorithm.


\subsection{Optimization Problem Formulation}
\label{ssec:optm_form}

For a complete path planning problem formulation as an optimization problem, the cost function and problem constraints must be defined.

\subsubsection{Cost Function}
Given the starting point (\textbf{S}) and the target point (\textbf{T}) defined as $S=P_1$ and $T=P_n$, the objective is to find a decision variables matrix, $\textbf{L}=[P_1,P_2,...,P_{n-1},P_n]$, such that, $J(L)$ is the length function. The cost function is defined by Eq.~\eqref{eq:costfunction} as
\begin{equation}
\label{eq:costfunction}
J(L) = \sum_{i=1}^{n-1} \left\| P_{i+1} - P_i \right\|_2,
\end{equation}
\noindent where $n$ is the number of points that compose the path and for the bi-dimensional case, $P_i=(x_i,y_i)$ is a path vertex.

We can see that the bi-dimensional path planning optimization problem is an optimization problem in $2n-4$-th dimension. Note that, if $n\rightarrow \infty$, the path will be a smooth trajectory; furthermore, the optimization problem dimension will also tend to infinity. However, the trajectory smoothness should be provided by a trajectory planning algorithm, which uses the results of the path planning algorithm (the scope of this paper).

\subsubsection{Constraints}

According to Definition~\ref{def:path}, the path is formed by $n-1$ straight segments, which connect the $n$ points such that the $i$-th straight segment is built from $P_i$ to $P_{i+1}$. Thus, constraints are about each point $p_{i\lambda}$ that composes the $i$-th straight segments, in such way that the $i$-th must not intercept any obstacle. From Definition~\ref{def:def1} and Eq.~\eqref{eq:costfunction}, the path planning optimization problem can be written as
\begin{equation}
\label{eq:pathproblem}
\begin{array}{cc}
\min\limits_{L} & J(L),  \\
 & p_{i\lambda}(L) \notin \mathbb{O} \\
\textrm{ s.t. } & p_{i\lambda}(L) \in \mathbb{E} \\
& i=1, ..., n-1, \\
\end{array}
\end{equation}
\noindent where $\mathbb{O}$ is the set of points defined by obstacles; $\mathbb{E}$ is the set of points defined by environment limits; $n$ is the number of points that compose the path; and $p_{i\lambda}(L)$ is all points belonging to the $i$-th straight segment of the path defined by vector \textbf{L}, each $p_{i\lambda}(L)$ point is defined by Eq.~\eqref{eq:points} as
\begin{equation}
\label{eq:points}
p_{i\lambda}(L) = (1-\lambda) P_i + \lambda P_{i+1}, \forall \lambda \in [0,1].
\end{equation}

\subsubsection{Environmental Modeling}

Although the optimization problem is defined by Eq.~\eqref{eq:pathproblem}, it is still necessary to model the environment of movement ($\mathbb{E}$) and obstacles ($\mathbb{O}$) in it. For simplicity, the movement environment is modeled as a rectangle, which is defined by lower and upper limits and constrains the value that each coordinate can assume.

Obstacles can be modeled by means of circles, such that the geometric center of the real obstacle shape defines the center of the circle; the circle radius is such that every point of the real obstacle shape is inside the circle. The constraints of the optimization problem~\eqref{eq:pathproblem} ensure that there is no intersection between the path segments and the obstacle. This constrained is described by Eq.~\eqref{eq:obstacles}.
\begin{equation}
\label{eq:obstacles}
(x_{i\lambda}-x_0)^2+(y_{i\lambda}-y_0)^2 \geq (r+\sigma)^2
\end{equation}
where $p_{i\lambda}=(x_{i\lambda},y_{i\lambda})$, and $\sigma$ is a safety margin.

\subsection{Path Planning Algorithm}
\label{ssec:path_algo}

There are two code directive in C/C++ programming language, which can be used to model and control the verification process: ASSUME and ASSERT. The ASSUME directive is able to define constraints over (non-deterministic) variables, and the ASSERT directive is used to check system correctness with respect to a given property. Using these two directives, any off-the-shelf C/C++ model checker could be applied to check specific constraints in optimization problems.

The verification process consists of three steps: modeling, especification, and verification \cite{Araujo2016}. Thus, the optimization problem described in section \ref{ssec:optm_form} is encoded as shown in Figure~\ref{fig:code_traj}. This C code checks whether the literal $J_{optimal}$ given by Eq.~\ref{eq:lotimo1} is satisfied for value $J_c$ that is a candidate to optimal, such that $J_c$ is randomly initialized with high values.
%
\begin{equation}
\label{eq:lotimo1}
\begin{array}{cc}
J_{optimal} \iff J(\textbf{L}) > J_c
\end{array}
\end{equation}

A function \texttt{rest\_points} inserts the constraints on $n-1$ path straight segments, in such way that they satisfy the conditions discussed in the previous section; for this purpose, the ASSUME directive is used. Note that \texttt{rest\_points} must be executed for each obstacle. Figure~\ref{fig:rest} illustrates the \texttt{rest\_points} ANSI-C code.
\begin{figure}[ht!]
	\scriptsize
	\begin{lstlisting}[xleftmargin=0.01\textwidth,xrightmargin=0.01\textwidth,frame=single]
#define DIM 2	// space dimension
#define n 1	// number of points that compose the path
#define p 1	// precision of points localization
#define J_c 25	// candidate value of cost function
#define no 1	// number of obstacles
// obstacles information
float x0[no] = {5};	// coordinates of center 'x'
float y0[no] = {5};     // coordinates of center 'y'
float r[no] = {2.5};    // obstacles radius
int main() {
   int i, j;
   int A [DIM] = {1*p, 1*p};	// start point
   int B [DIM] = {9*p, 9*p};	// target point
   // environmental limits
   int lim [DIM][2] = { 0*p, 10*p, 0*p, 10*p};
   // states declaration, x=x[i][0] and y=x[i][1] 
   // as non-deterministic
   for (i=0; i<n; i++)
     for(j=0; j<DIM; j++)
         x[i][j] = nondet_int();
   // constraints on environment limits and obstacles
   for (i=0; i<n; i++) {
      __ESBMC_assume( x[i][0] >= lim[0][0] );
      __ESBMC_assume( x[i][0] <= lim[0][1] );
      __ESBMC_assume( x[i][1] >= lim[1][0] );
      __ESBMC_assume( x[i][1] <= lim[1][1] );
   }
   for (j=0; j<no; j++)
     rest_points (A, 0, x0[j]*p, y0[j]*p, r[j]*p);
   for (i=1; i<n; i++) {
     for (j=0; j<no; j++)
       rest_points(x[i-1],i,x0[j]*p,y0[j]*p,r[j]*p);
   }
   for (j=0; j<no; j++)
     rest_points (B, n-1,x0[j]*p,y0[j]*p,r[j]*p);
   // compute the cost function
   float Aux1[DIM], Aux2[DIM];
   float J = 0.0;
   for (j=0; j<DIM; j++)	
     Aux1[j] = A[j]/p;
   for (i=0; i<n; i++) {
     for (j=0; j<DIM; j++)
       Aux2[j] = (float) x[i][j]/p;
     J = J + dist(Aux1,Aux2);
     for (j=0; j<DIM; j++)
       Aux1[j] = Aux2[j];
   }
   for (j=0; j<DIM; j++)
     Aux2[j] = B[j]/p;
   J = J + dist(Aux1,Aux2);
   __ESBMC_assume( J < J_c );
   // test the literal J_optimal, Eq. (7)
   assert ( J > J_c );
   return 0;
}
\end{lstlisting}
\caption{C code for bi-dimensional path planning.}
	\label{fig:code_traj}
\end{figure}

\begin{figure}[ht!]
	\scriptsize
	\begin{lstlisting}[xleftmargin=0.01\textwidth,xrightmargin=0.01\textwidth,frame=single]
void rest_points(int P1 [DIM], int i, float x0, float y0, 
                float r) {
   float sigma = 0.5;	// safety margin
   // constraint given by Eq. (6)
   __ESBMC_assume( (x[i][0]-x0)*(x[i][0]-x0) +
   (x[i][1]-y0)*(x[i][1]-y0) > (r+sigma)*(r+sigma) );
   float a, b, c;
   if (P1[0]-x[i][0]==0){
      a = 1;
      b = 0;
      c = -P1[0];
   }
   else{
      a = (float) (P1[1]-x[i][1])/(P1[0]-x[i][0]);
      b = -1;
      c = (float) -a*P1[0]+P1[1];
   }
   float Py = (a*a*y0-a*b*x0-b*c)/(a*a+b*b);
   if (((Py-x[i][1])/(P1[1]-x[i][1])>=0) &&
      ((Py-x[i][1])/(P1[1]-x[i][1])<=1))) {
      float d=(float) abs2(a*x0+b*y0+c)/sqrt2(a*a+b*b);
      __ESBMC_assume( d > r );
   }
}
	\end{lstlisting}
	\caption{C code for function \texttt{rest\_points}.}
	\label{fig:rest}
\end{figure}

If the code shown in Figure~\ref{fig:code_traj} returns false, \textit{i.e.}, the negation of $J_{optimal}$ is satisfiable, then there is a $\textbf{L}^{(i)}$ for which $J(\textbf{L}^{(i)})<J_c$. Thus, the optimal candidate, $J_{c}$, can be updated with the returned value, $J_c=J(\textbf{L}^{(i)})$, for the new code instance execution; otherwise, $J_{optimal}$ is unsatisfiable, {\it i.e.}, $J(\textbf{L}^{(i-1)})$ is the optimal value for a given precision $p$. Furthermore, the counterexample returns the matrix \textbf{L} that defines the optimal path with $n$ points between start and target points.

The number of points is automatically increased if it is not possible to find an optimal path, \textit{i.e.}, $J_{optimal}$ is unsatisfiable. If $J_{optimal}$ is repeatedly unsatisfiable, then the precision must be improved. The precision $p$ defines the path points coordinate precision, such that:
\begin{equation}
\label{eq:precision}
k > log p,
\end{equation}
\noindent where $k$ is the number of decimal places of the points coordinate values. The precision $p$ is initialized by one, \textit{i.e.}, $k=0$ and coordinates are considered to be integers. The precision is increased by multiplying $p$ by $10$, {\it i.e.}, by adding one decimal place in the coordinate values. 

The optimization problem~\eqref{eq:pathproblem} is solved by executing the code in Figure \ref{fig:code_traj} iteratively, according to precision values and number of points that compose the path. Algorithm~\ref{alg:traj_optalg} summarizes the aforementioned steps  of the proposed bi-dimensional path planning methodology. The algorithm is inspired by CEGIO-F algorithm presented by Ara\'ujo {\it et al.}~\cite{Araujo2017}, since the cost function is convex.
\begin{algorithm}[h!]
\scriptsize
	\SetAlgoLined
	\LinesNumbered
	\SetKwData{Left}{left}\SetKwData{This}{this}\SetKwData{Up}{up}
	\SetKwFunction{Union}{Union}\SetKwFunction{FindCompress}{FindCompress}
	\SetKwInOut{Input}{input}\SetKwInOut{Output}{output}
	
	\Input{Cost function $J(\textbf{L})$, is a set of obstacles constraints $\mathbb{O}$ and a set of environment constraints $\mathbb{E}$, which define $\Omega$ and a desired precision $\eta$}
	\Output{The optimal path $\textbf{L}^{*}$ and the optimal cost function value $J(\textbf{L}^{*})$}
	
	\BlankLine
	\emph{Initialize $J(\textbf{L}^{(0)})$ randomly}\;
	\emph{Initialize precision variable with $p=1$, $k=0$ e $i=1$}\;
	\emph{Initialize number of points, $n=1$}\;
	\emph{Declare decision variables vector $\textbf{L}^{i}$ as non-deterministic integer variables}\;
	\While{$k\leq \eta$}{
		\emph{Define upper and lower limits of $\textbf{L}$ with directive \texttt{ASSUME}, such as $L\in \Omega^k$}\;
		\emph{Describe the objective function model $J(\textbf{L})$}\;
		
		\Do{TRUE}{
			\Do{$\neg J_{optimal}$ is satisfiable}{
				\emph{Define the constraint $J(\textbf{L}^{(i)})<J(\textbf{L}^{(i-1)})$ with directive \texttt{ASSUME}}\;
				\emph{Verify the satisfiability of $J_{optimal}$ given by Eq.~\eqref{eq:lotimo1}}\;
				\emph{Update $\textbf{L}^{*}=\textbf{L}^{(i)}$ e $J(\textbf{L}^{*})=J(\textbf{L}^{(i)})$ based on the counterexample}\;
				\emph{Do $i=i+1$}\;
			}
			\If {$\neg J_{optimal}$ is not consecutively satisfiable}{break}
			\Else{\emph{Update the number of points, $n$}\;}
		}
		\emph{Do $k=k+1$}\;
		\emph{Update the set $\Omega^k$}\;
		\emph{Update the precision variable, $p$}\;
	}
	\emph{$\textbf{L}^{*}=\textbf{L}^{(i)}$ e $J(\textbf{L}^{*})=J(\textbf{L}^{(i)})$}\;
	\Return{$\textbf{L}^{*}$ e $J(\textbf{L}^{*})$}\;
	\caption{Path planning algorithm based on satisfiability.}
	\label{alg:traj_optalg}
\end{algorithm}

The algorithm efficiency depends on the number of points in the path $n$. Naturally, a large value of $n$ will generate smooth paths, but it will increase the complexity of the optimization problem, leading to a large execution time. The smoothness is not required from the path planning algorithm, but it is provided by a trajectory planning algorithm that computes the curves between the points of the path, {\it i.e.}, a trajectory planning algorithm is responsible for interpolating the points found by the path planning.

\section{Experimental Evaluation}
\label{sec:exps}

\subsection{Experimental Objectives and Description}
\label{ssec:expobj}

The path planning algorithm described in section~\ref{sec:method} is suitable for a general autonomous vehicle, and the goal is to find points which compose the path. Two experiments were designed and executed to evaluate the application of the CEGIO-based path planning algorithm for an autonomous vehicle.

\begin{figure}[ht]
	\centering
	\subfigure[setting1][Setting 1]{\includegraphics[width=0.49\columnwidth]{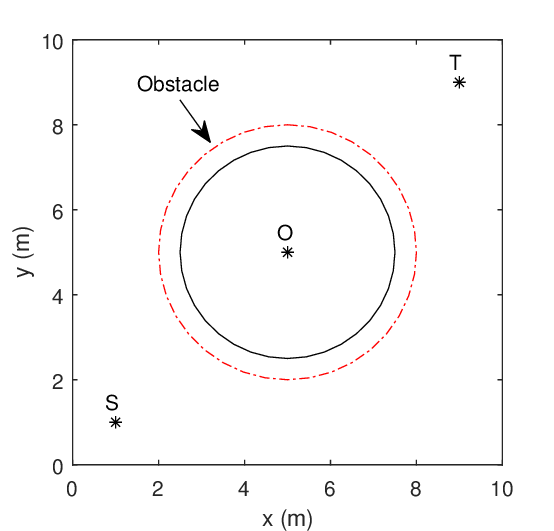}}
	\subfigure[setting2][Setting 2]{\includegraphics[width=0.49\columnwidth]{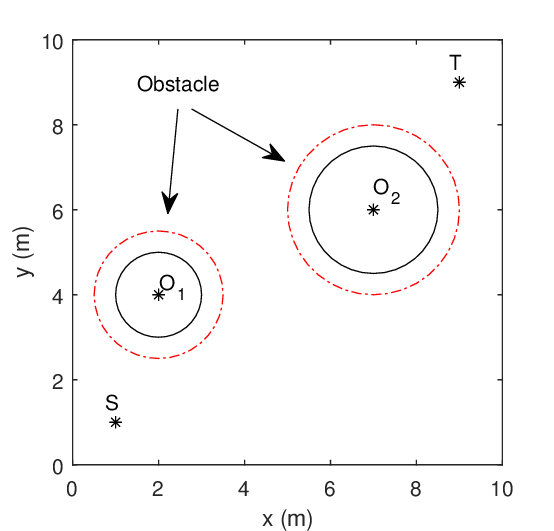}}
	\caption{Settings for evaluating the proposed Algorithm~\ref{alg:traj_optalg}.}
	\label{fig:setting}
\end{figure}

The first experiment is performed with Setting 1, in which the autonomous vehicle is inserted in a bi-dimensional space with the goal of avoiding a single obstacle, as shown in Figure~\ref{fig:setting}(a). The second experiment, with Setting 2, is similar to the first one, except that now there are two obstacles, as illustrated in Figure~\ref{fig:setting}(b).

The objective in both experiments is to compute a path from \textbf{S} to \textbf{T}. The precision on the points location that form the path is $10 cm$. 
In Setting 1, the obstacle is centered in $O(x_0,y_0)=(5,5)$ and its radius is $r=2.5$. In Setting 2, the obstacles are centered in $O_1(x_0,y_0)=(2,4)$ and $O_2(x_0,y_0)=(7,8)$, and their radius are $r_1=1$ $r_2=1.5$, respectively. The safety margin for both settings is $\sigma=0.5$. All previous distances are measured in meters. The maximum algorithm execution time for both employed verifiers CBMC\footnote{http://www.cprover.org/cbmc/} and ESBMC\footnote{http://esbmc.org/} is set to three days for Setting 1 and one week for Setting 2.


\subsection{Experimental Setup}
\label{ssec:expenv}

All experiments were conducted on an otherwise idle Intel Core i$7-4790$ $3.60$ GHz processor, with $16$ GB of RAM, running Ubuntu 14.10 $64$-bits. Additionally, the time presented here is related to the average of $10$ executions for each benchmark; the measuring unit is always in seconds based on the CPU time; we did not restrict the memory consumption for the experiments.
The employed software verifiers versions are: CBMC v4.5 with support to the MiniSAT v$2.2.0$ solver and ESBMC v$3.1.0$ with support to the MathSAT v$5.3.13$ solver.

\subsection{Experimental Results}
\label{ssec:expres}

\begin{figure}[ht]
	\centering
	\subfigure[path1][Path generated for Setting 1]{\includegraphics[width=0.49\columnwidth]{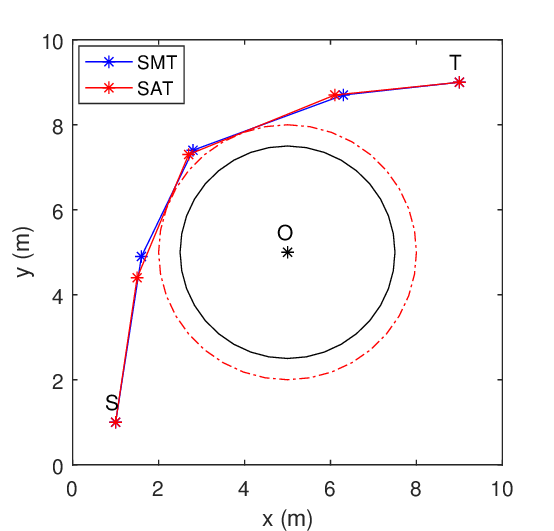}}
	\subfigure[path2][Path generated for Setting 2]{\includegraphics[width=0.49\columnwidth]{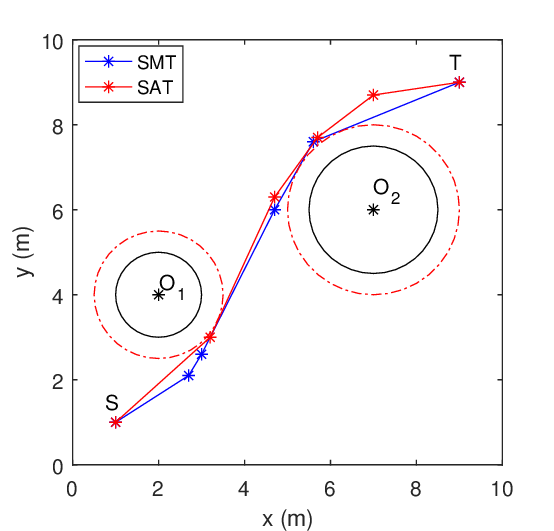}}
	\caption{Paths obtained by Algorithm \ref{alg:traj_optalg}. Step of the cost function is $10^{-4}$.}
	\label{fig:paths}
\end{figure}

The paths obtained in Algorithm~\ref{alg:traj_optalg} for both settings and also using the SAT and SMT solvers are illustrated in Figure~\ref{fig:paths}. For Setting 1, a path with only five points was obtained ($n=5$), and for Setting 2, the obtained path has six points ($n=6$); both scenarios suffered the timeout of three days and one week, respectively.

\begin{figure}[ht]
	\centering
	\subfigure[traj1][Trajectory for Setting 1]{\includegraphics[width=0.49\columnwidth]{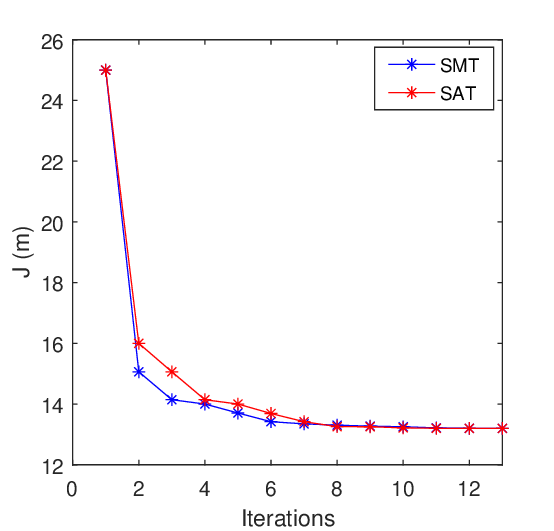}}
	\subfigure[traj2][Trajectory for Setting 2]{\includegraphics[width=0.49\columnwidth]{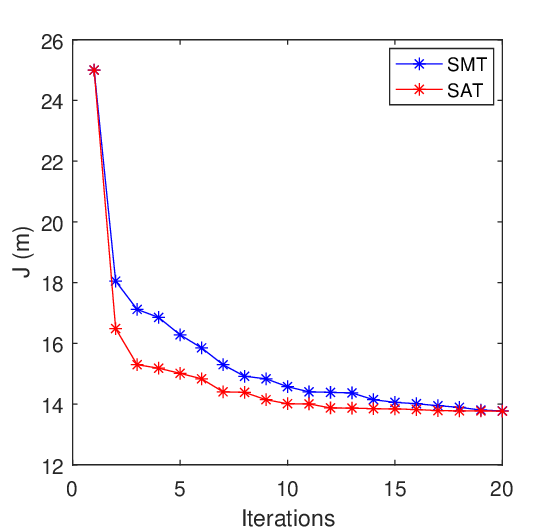}}
	\caption{Paths obtained by Algorithm \ref{alg:traj_optalg}. Step of the cost function is $10^{-4}$.}
	\label{fig:trajs}
\end{figure}

Figure~\ref{fig:trajs} shows the cost function convergence trajectories to obtain those paths shown in Figure~\ref{fig:paths}. Note that the cost function value always decreases at each iteration and converges to the optimal solution. However, it does not reach the optimal value due to the timeout previously defined. SAT and SMT solvers obtained very similar solutions with the same number of points and even timeout; however, the SAT solver converges more quickly to the optimal solution if the number of points increases, as shown in Figure~\ref{fig:trajs}(b).

The timeout for both settings occurred due to the cost function step being $10^{-4}$, which is the precision of values returned from the counterexamples. Thus, the proposed algorithm requires more iterations to converge to the optimal solution and as much close to that, each iteration does not significantly improve the value of the cost function, which substantially increases the convergence time.

A solution to this problem is to increase this step, whereby the value of the cost function decreases. Thus, in order to evaluate the step influence in the execution time, the step was fixed to $10^{-2}$, \textit{i.e.}, $J_c=J(\textbf{L}^{(i)})-10^{-2}$, a hundred times larger than the previous step, and the Setting 1 experiment is repeated. However, the found solution will not be the best, but it will be at a distance of $10^{-2}$ from it and for most problems, it is still a satisfactory solution.

\begin{figure}[ht]
	\centering
	\subfigure[pathcomp][Paths for Setting 1]{\includegraphics[width=0.49\columnwidth]{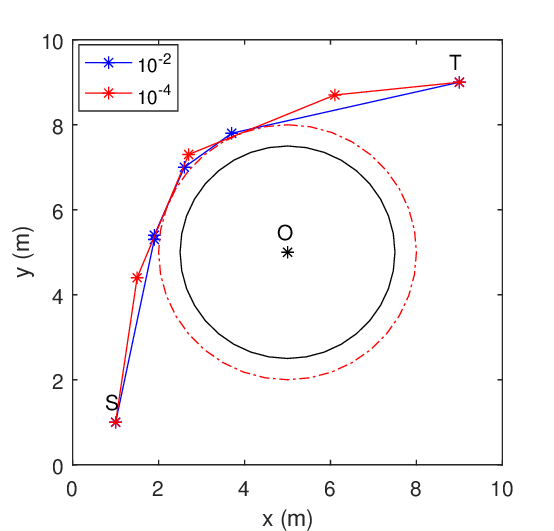}}
	\subfigure[trajcomp][Trajectories for Setting 1]{\includegraphics[width=0.49\columnwidth]{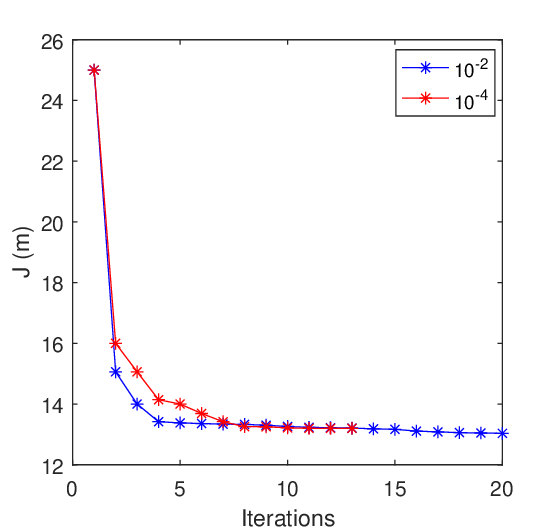}}
	\caption{Comparison of cost function paths and trajectories for steps $10^{-2}$ and $10^{-4}$.}
	\label{fig:comparison}
\end{figure}

Figure~\ref{fig:comparison} shows a comparison between the obtained paths and the convergence trajectory of cost function for Setting 1, considering the steps previously mentioned, $10^{-4}$, and the step fixed in $10^{-2}$. Only the SAT solver was used to perform this specific experiment.

The obtained path, considering the new step, has more points, $n=6$; additionally, it was found in a much shorter time than the previous configuration, {\it i.e.}, only four hours, which represents $5.5\%$ of the previous time, $72$ hours, although it took more iterations. This step can be further increased such that satisfactory solutions can be obtained in a shorter time. However, there is no guarantee that the optimal solution is found, as previously mentioned, it is only guaranteed that the solution found is at a distance relative to the chosen step of the optimal solution.



\section{Conclusion}
\label{sec:conc}
We have presented a novel path planning algorithm for mobile robots, where the optimal path planning problem is  solved by the application of the CEGIO-F algorithm. The experimental results indicate that CEGIO is able to provide optimal paths for mobile robots. However, the cumulative execution time is still high, if compared to traditional optimization-based path planning algorithms. Future studies consist in applying the CEGIO-based algorithm to tridimensional environments and in the context of the trajectory planning in order to obtain a smooth trajectory.

\end{document}